\def\year{2021}\relax
\documentclass[letterpaper]{article} 
\usepackage{aaai21}  
\usepackage{times}  
\usepackage{helvet} 
\usepackage{courier}  
\usepackage[hyphens]{url}  
\usepackage{graphicx} 
\usepackage{graphicx}  
\usepackage{natbib}  
\usepackage{caption} 
\usepackage{amsmath}
\usepackage{amssymb}
\usepackage[ruled,vlined]{algorithm2e}
\SetKwInput{KwInput}{Input}                
\SetKwInput{KwOutput}{Output}              

\usepackage{array}
\usepackage{multirow}

\usepackage{booktabs}
\usepackage{subcaption}
\usepackage{pifont}
\usepackage{color}

\title{Feature Distillation With Guided Adversarial Contrastive Learning}
\author{
    Tao Bai,
    Jinnan Chen,
    Jun Zhao,
    Bihan Wen,
    Xudong Jiang,
    Alex Kot
}
\affiliations{

    \textsuperscript{\rm 1}Nanyang Technological University\\

    \{bait0002, jinnan001, junzhao, bihan.wen, exdjiang, eackot\}@ntu.edu.sg

}

\begin{document}
\maketitle

\begin{abstract}

Deep learning models are shown to be vulnerable to adversarial examples.
Though adversarial training can enhance model robustness, typical approaches are computationally expensive.
Recent works proposed to transfer the robustness to adversarial attacks across different tasks or models with soft labels.
Compared to soft labels, feature contains rich semantic information and holds the potential to be applied to different downstream tasks.
In this paper, we propose a novel approach called Guided Adversarial Contrastive Distillation (GACD), to effectively transfer adversarial robustness from teacher to student with features.
We first formulate this objective as contrastive learning and connect it with mutual information.
With a well-trained teacher model as an anchor, students are expected to extract features similar to the teacher.
Then considering the potential errors made by teachers, we propose sample reweighted estimation to eliminate the negative effects from teachers.
With GACD, the student not only learns to extract robust features, but also captures structural knowledge from the teacher.
By extensive experiments evaluating over popular datasets such as \mbox{CIFAR-10}, CIFAR-100 and STL-10, we demonstrate that our approach can effectively transfer robustness across different models and even different tasks, and achieve comparable or better results than existing methods.
Besides, we provide a detailed analysis of various methods, showing that students produced by our approach capture more structural knowledge from teachers and learn more robust features under adversarial attacks.


\end{abstract}

\section{Introduction}
Deep neural networks (DNN) have achieved impressive performances for many computer vision~\cite{lecun2015deep,he2016deep} and natural language processing~\cite{mikolov2013distributed} tasks. 
Nevertheless, they are vulnerable to adversarial examples~\cite{DBLP:journals/corr/SzegedyZSBEGF13}, and their performances degrade sharply. 
Such adversarial examples are usually crafted by adding human-imperceptible perturbations, and able to mislead well-trained models at inference time.
Due to the potential safety risks, the existence of adversarial examples has been a crucial threat to safety and reliability critical applications~\cite{parkhi2015deep,liu2019high}. Thus, extensive researchers have devoted to studying and enhancing the robustness of DNNs~\cite{schott2018towards,Zhong_2019_ICCV,DBLP:conf/cvpr/PrakashMGDS18,NIPS2019_8339,DBLP:conf/iclr/MadryMSTV18} against waves of new adversarial attacks~\cite{Papernot1602,carlini2017towards,Croce_2019_ICCV,modas2019sparsefool}, where adversarial training~\cite{DBLP:journals/corr/GoodfellowSS14,DBLP:journals/corr/abs-1803-06373,DBLP:conf/iclr/MadryMSTV18,NIPS2019_8597} generates models with best adversarial robustness, and undoubtedly is recognized as the strongest defense method. 

Adversarial training is based on a simple yet effective idea.
Compared to traditional training, it only requires to involve training models with adversarial samples generated in each training loop.
It, however, is computationally expensive and time-consuming as it needs multiple gradient computations to craft strong adversarial samples~\cite{NIPS2019_8597}.
To circumvent the high cost of adversarial training, recently various methods have been developed and introduced to transfer adversarial robustness between models~\cite{papernot2016distillation,DBLP:conf/aaai/GoldblumFFG20}. 
Currently existing methods are based on the idea of distillation~\cite{hinton2015distilling}, and use soft labels. 

Another problem neglected in distillation for adversarial robustness is that there are many errors made by the teachers. 
Compared to the naturally trained models, adversarially trained models usually have a performance degradation on classification accuracy, which is theoretically proved in~\cite{tsipras2018robustness,DBLP:conf/eccv/SuZCYCG18,pmlr-v97-zhang19p}.
In existing distillation methods, errors from teachers are transferred to students as well.

Here, we propose a teacher-error aware method to transfer robustness with features, called Guided Adversarial Contrastive Distillation (GACD). 
Our method is inspired by the observation that features learned by DNNs are purified during adversarial training (see Figure 5 in~~\cite{2020arXiv200510190A}), which are more aligned with input images in semantic.
\cite{DBLP:conf/cvpr/XieWMYH19} empirically proved models with purified features are more robust to adversarial examples.
So the core idea of GACD is training a student model to imitate the teacher model to extract similar robust features under some similarity metrics. 
With the context of contrastive learning, the teacher model is used as an anchor and the student model learns to be aligned with the the teacher.
However, the anchor is not always reliable as explained in last paragraph.
So we use sample re-weighting to eliminate the negative effects caused by bad anchorS.
Through experiments in \mbox{CIFAR-10}, \mbox{CIFAR-100} and \mbox{STL-10}, we show that our approach outperforms other methods in classification accuracy and transferability.
Some students are even better than the teachers.




In summary, the key contributions of this paper are as follows:
\begin{itemize}
  \item To our best knowledge, we are the first to transfer adversarial robustness across different model architectures with features. Based on contrastive learning, we proposed a novel approach called GACD with awareness of teacher's mistakes.
  \item Through extensive experiments on CIFAR-10, CIFAR-100 and STL-10, it is shown that our method outperforms existing distillation methods, achieving comparable or higher classification accuracy and better transferability.
  \item By conducting analysis of different methods in feature space, we show our method is more effective in capturing structural knowledge from teachers and extracts more gathered and robust features between classes.
\end{itemize}





\section{Related Work}
In this section, we review the prior work on adversarial attacks, adversarial training and robustness distillation.
\subsubsection{Adversarial Training.}
Since~\cite{DBLP:journals/corr/SzegedyZSBEGF13} first proved the existence of adversarial examples, extensive adversarial attacks~\cite{papernot2016limitations,DBLP:conf/cvpr/DongLPS0HL18,carlini2017towards,DBLP:conf/iclr/MadryMSTV18} have been proposed, showing the vulnerability of neural networks facing inputs with imperceptible perturbations.
\cite{DBLP:journals/corr/GoodfellowSS14} assumed the vulnerability of neural networks caused by linearity and proposed a single-step method called Fast Gradient Sign Methods~(FGSM). 
Then iterative attacks~\cite{DBLP:journals/corr/KurakinGB16,DBLP:conf/iclr/MadryMSTV18} were studied and proved to be stronger than FGSM under the same metric norm.
Momentum is introduced in~\cite{DBLP:conf/cvpr/DongLPS0HL18} and enhanced the transferability of untargeted attacks.


As adversaries could launch attacks with 100\% success rate on well-trained models with small perturbation budget $\epsilon$, defense methods~\cite{DBLP:conf/iclr/MadryMSTV18,DBLP:conf/cvpr/LiaoLDPH018,DBLP:conf/cvpr/PrakashMGDS18,2020arXiv200207405Q,Gupta_2019_ICCV} are developed right after.
Adversarial training is proved to be most effective among these methods, which simply requires to train models on adversarial examples progressively. 
Intuitively, adversarial training encourages models to predict correctly in an \mbox{$\epsilon$-ball} surrounding data points. 
Many variants of adversarial training have been developed from this observation. 
Recently \cite{NIPS2019_8459} introduced a feature scatter-based approach for adversarial training, which generates adversarial examples in latent space in a unsupervised way.
\cite{NIPS2019_9534} noticed the highly convoluted loss surface by gradient obfuscation~\cite{pmlr-v80-athalye18a}, and introduced a \textit{local linearity regularizer (LLR)} to adversarial training to encourage the loss surface behave linearly.
\cite{pmlr-v97-zhang19p} decomposed the classification errors on adversarial examples and proposed a regularization term to improve adversarial robustness.

\subsubsection{Self-Supervised Learning.}
Based on deep learning models, self-supervised learning methods are becoming more and more powerful, and gaining increasing popularity.
The normal approach of self-supervised learning is training models to learn general representations out of unlabeled data, which can be later used for specific tasks, like image classification.
Predictive approaches have shown to be effective to learn representations~\cite{doersch2015unsupervised,zhang2017split,noroozi2016unsupervised}.
Contrastive learning is another collection of powerful methods for self-supervised representation learning.
Inspired by~\cite{gutmann2010noise,mnih2013learning,sohn2016improved}, the core idea of contrastive learning is learning representations that is close under some distance metrics for samples in same classes (positive samples) and pushing apart representations between different classes (negative samples).
By leveraging the instance-level identity for self-supervised learning, contrastive learning are shown to be effective in learning representations~\cite{2020arXiv200205709C,2019arXiv191105722H,tian2020makes}, and achieves comparable performances to supervised methods. 
Note that such methods are used to learn embeddings. 
At test time, the embeddings are utilized for other tasks with fine tuning.

\subsubsection{Knowledge Distillation.}
Distillation is originally introduced in~\cite{hinton2015distilling} to compress model size while preserving performances with a student-teacher scheme.
The student usually has a small, lightweight architecture.
Thereafter, more distillation methods are proposed~\cite{heo2019comprehensive,zhang2019your,DBLP:conf/cvpr/ParkKLC19,Tian2020Contrastive}.
\cite{DBLP:conf/cvpr/ParkKLC19} noticed relationships between samples are usually neglected and proposed \textit{Relational Knowledge Distillation (RKD)}, while~\cite{Tian2020Contrastive} adapted contrastive learning into knowledge distillation and formulated the distillation problems as maximizing the mutual infomation between student and teacher representations.
And \cite{zhang2019your} proposed self distillation to distill knowledge from the model itself.
Note that such distillation methods only preserve performances in student models. 

Knowledge distillation is firstly adapted for obtaining or transferring adversarial robustness in~\cite{papernot2016distillation}, which is called \textit{defensive distillation}.
Defensive distillation requires student and teacher models have the identical architectures. Due to gradient masking, defensive distillation improves the robustness of the student model under a certain attack. It, however, doesn't make the decision boundary secure and is circumvented by~\cite{carlini2017towards} soon. 
Recently \cite{DBLP:conf/aaai/GoldblumFFG20} studied how to distill adversarial robustness onto student models with knowledge distillation.
And a line of work explores transferring adversarial robustness between models with same architectures.
\cite{DBLP:conf/icml/HendrycksLM19} shows large gains on adversarial robustness from pre-training on data from different domain.
Another work uses the idea of transfer learning and builds new models on the top of robust feature extractors~\cite{shafahi2020adversarially}.
In contrastive to these methods, our method focuses on robust feature distillation, and is not restricted to same model architectures. 

\section{Guided Adversarial Contrastive Distillation}
In this section, we start by defining feature distillation between teachers and students. 
Then we formulate the problem with contrastive learning and show the connection between our objective and mutual information.
Lastly we derive our final objective function.

\subsection{Problem Definition}
\begin{figure}[h]
\centering
\includegraphics[width=\linewidth]{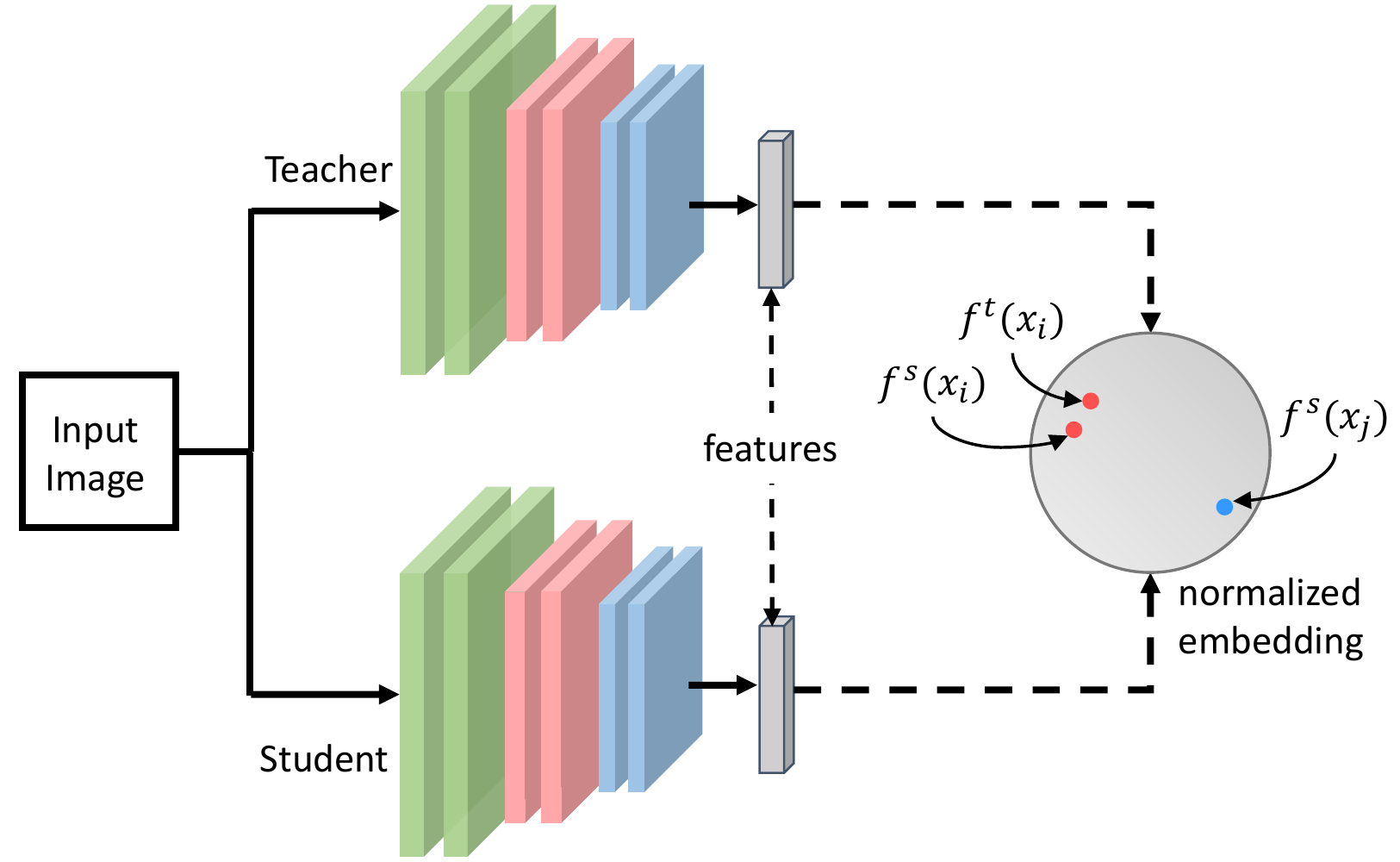}
\caption{Illustration of feature distillation with contrastive learning.}
\label{fig:overview}
\end{figure}

Teacher-student paradigm is widely used in knowledge distillation. 
Given two deep neural networks, a teacher $f^{t}$ and a student $f^{s}$, and their architectures are not necessarily to be identical.
Let $\{(x, y)  \mid  x \in \mathcal{X}\}$ with $K$ classes be the training data. Then the representations extracted at the penultimate layer (the layer before logits) are denoted as $f^{t}(x)$ and $f^{s}(x)$. 
During distillation, for two random samples $x_i$ and $x_j$, we expect to push representations $f^{s}(x_i)$ and $f^{t}(x_i)$ closer if $i=j$, while pushing $f^{s}(x_i)$ and $f^{t}(x_j)$ apart if $i\neq j$. 
Figure~\ref{fig:overview} gives a visual explanation of this intuition.
Note that we use adversarially robust models as teacher models as in~\cite{DBLP:conf/aaai/GoldblumFFG20}.
The features extracted by the teacher and student are transformed to the same dimension, refering to normalized embedding.

\subsection{Connecting to Mutual Information}
Contrastive learning is commonly used to extract features in a unsupervised way, and the core idea is learning representations that is close under some distance metrics for samples in same classes (positive samples) and pushing apart samples from different classes (negative samples).
Following the recent setups for contrastive learning~\cite{gutmann2010noise,mnih2013learning,tian2019contrastive}, we select samples from training data $\mathcal{X}: \left\{x \sim p_{\text {data }}(x)\right\}$, and construct a set $\mathcal{S}=\left\{x^{+},x^{-}_{1},x^{-}_{2} \dots x^{-}_{k}\right\}$ where there are a single positive samples and $k$ negative samples from different classes.

Since the teacher model is pretrained and fixed during distillation, we simply enumerate positives and negatives for student model from $\mathcal{S}$.
Then we have one contrastive congruent pair (the same input given to teacher and student model) for every $k$ incongruent pairs (different inputs given to teacher and student model) as given in 
$\mathcal{S}_{pair}:=\{(t^{+},s^{+}),(t^{+},s^{-}_{1}),(t^{+},s^{-}_{2}) \dots (t^{+},s^{-}_{k})\}$,
where we denote $f^{t}(x)$ as $t$, and $f^{s}(x)$ as $s$ for simplicity.

Now we define a distribution $q$ with latent variable $C$:
\begin{itemize}
    \item when $C=1$, the congruent pair $(t, s)$ is drawn from the joint distribution $p(t,s)$, 
    \begin{equation}q(t,s \mid C=1)=p(t,s) .\label{eqn:q c=1}\end{equation}
    \item when $C=0$, the incongruent pair $(t, s)$ is drawn from the joint distribution $p(t)p(s)$, 
    \begin{equation}q(t,s \mid C=0)=p(t)p(s) .\end{equation} 
\end{itemize}
According to set $S$, the priors on $C$ are
\begin{equation}q(C=1)=\frac{1}{k+1}, \quad q(C=0)=\frac{k}{k+1}.
\label{eqn:prior}
\end{equation}
By applying Bayes' rule, we can easily derive the posterior for class $C=1$:
\begin{equation}\begin{aligned}
&\quad q(C=1 \mid t, s) \\ &= \frac{q(t, s \mid C=1) q(C=1)}{q(t, s \mid C=0) q(C=0)+q(t, s \mid C=1) q(C=1)} \\
&=\frac{p(t, s)}{k p(t) p(s)+p(t, s)} \leq \frac{p(t, s)}{k p(t) p(s)}.
\end{aligned}
\label{eqn:posterior c=1}
\end{equation}
Then taking log of both sides of (\ref{eqn:posterior c=1}), we have
\begin{equation}\begin{aligned}
\log q(C=1 \mid t, s) &\leq \log \frac{p(t, s)}{k p(t) p(s)} \\
&\leq-\log (k)+\log \frac{p(t, s)}{p(t) p(s)}.
\end{aligned}\end{equation}

By taking expectation on both sides w.r.t. $p(t,s)$ or \mbox{$q(t, s \mid C=1)$} (they are equal as shown in Eq.~(\ref{eqn:q c=1})), the connection with mutual information is given as:
\begin{equation}MI(t ; s) \geq \log (k)+\mathbb{E}_{q(t, s \mid C=1)} \log q(C=1 \mid t, s),
\label{eqn:mutual info ori}
\end{equation}
where $MI(t ; s)$ represents the mutual information between $t$ and $s$.

Thus, our objective is to maximize the lower bound of mutual information, and consistent with~\cite{DBLP:journals/corr/abs-1807-03748,poole2019variational,tian2019contrastive,Tian2020Contrastive}.

\subsection{Sample Reweighted Noise Contrastive Estimation}
To maximize the right side of Inequality~(\ref{eqn:mutual info ori}), the distribution $q(C=1 \mid t, s)$ is required.
Though we don't have the true distribution, we can estimate it by fitting a model~\cite{gutmann2010noise,goodfellow2014generative} based on teacher-student pairs in $\mathcal{S}_{pair}$.

Before proceeding, we first retrospect the pair set.
As shown in $\mathcal{S}_{pair}$, the teacher is fixed and plays as an anchor.
If the anchor gets wrong, the whole set is not reliable.
And it is known that there is a large drop on performances of adversarially trained models as explained in~\cite{pmlr-v97-zhang19p,DBLP:conf/iclr/MadryMSTV18}.
The robust teacher models in adversarial settings are not as good as the naturally trained teacher models in benign settings (e.g. roughly 76\% v.s. 95\% on CIFAR-10).
Thus, a large amount of errors made by teacher models are transferred to student models along with knowledge during distillation, which is indeed serious but neglected in existing methods~\cite{shafahi2020adversarially,DBLP:conf/aaai/GoldblumFFG20}.

The direct way to handle this problem is removing samples which are misclassified by the teacher during training.
However, this will reduce the amount of training data.
Besides, such surrogate loss holds inherent limitations, such as computational hardness.
Instead, we propose Sample Reweighted Noise Contrastive Estimation, which assigns smaller weights to misclassified samples and greater weights to those classified correctly by the teacher~(see Algorithm~\ref{algo}).
With softmax outputs of teacher models at hand, we pick up the probability of true class for each sample as the weight, and denote it as $w_t$.
The higher the probability, the greater the chance of samples classified correctly.
Hence, the misclassified samples do not significantly affect the estimation process.

Now we formulate the estimation of $q(C=1 \mid t, s)$ as a binary classification problem and our goal is to maximize the log likelihood.
We use $h$ to represent the classification model, which takes $t$ and $s$ as inputs and gives the probability of $C$.
Then we have
\begin{equation}q(C=1 \mid t, s)=h(t,s), \end{equation}
\begin{equation}q(C=0 \mid t, s)=1-h(t,s).\end{equation}
The probabilities for two classes are given in Eq.~(\ref{eqn:prior}).
Considering the sample weights, the log-likelihood on $\mathcal{S}_{pair}$ is re-formulated as
\begin{equation}\begin{aligned}
\ell(h, w_t)=& \sum_{(t, s) \in \mathcal{S}_{pair}} w_t C\log  P\left(C=1 \mid t,s \right)+\\
&w_t\left(1-C\right) \log P\left(C=0 \mid t,s \right) \\
=& w_t\log \left[h\left(t^+,s^+ \right)\right]+ \sum_{i=1}^k w_t\log \left[1-h\left(t^{+},s^{-}_i \right)\right].
\end{aligned}\end{equation}
Formally the log likelihood of $h$ is expressed as
\begin{equation}\begin{aligned}
\mathcal{L}(h, w_t)= & \mathbb{E}_{q(t,s \mid C=1)}[w_t \log h(t,s)]+ \\
& k \mathbb{E}_{q(t,s \mid C=0)}[w_t \log (1-h(t,s))],
\end{aligned}
\label{eqn:loss h}
\end{equation}
where $\mathbb{E}_{q(t,s \mid C=0)}[w_t \log (1-h(t,s))]$ is strictly negative, and $w_t$ can not be larger than 1. 
Thus, adding the second term to the right side of Inequality~(\ref{eqn:mutual info ori}), it still holds.
We rewrite the inequality as below:
\begin{equation}\begin{aligned}
I(t ; s) & \geq  \log (k) + \mathcal{L}(h^*),
\label{eqn:mutual info final}
\end{aligned}
\end{equation}
where $\mathcal{L}(h^*)$ is the upper bound of $\mathcal{L}(h)$. 

To summarize, our final learning problem is to learn a student model $f^s$ to maximize the log likelihood $\mathcal{L}(h)$ (see Eq.~\ref{eqn:loss h}).


\begin{algorithm}[t]
\SetAlgoLined
\KwInput{Teacher model $f^t$, student model $f^s$, estimation model $h$, number of negatives $k$, learning rate $r$}
\KwOutput{The final student model parameter $\theta_s$}
\KwData{Training Data $\mathcal{D}_{train}$}
\For{each training iteration}{
    Sample $(x, y) \sim \mathcal{D}_{\text {train }}$ \;
    Construct $\mathcal{S}=\left\{x^{+},x^{-}_{1},x^{-}_{2} \dots x^{-}_{k}\right\}$ \;
    $t^{+} \leftarrow f^t(x^{+})$ \;
    $w_t \leftarrow p_{y_{pred = y}}(f^t, t^+)$ \;
    \For{$x^{-}_{i} \in \mathcal{S}$}{
        $s^{-}_{i} \leftarrow f^s(x^{+})$ \;
    }
    $\mathcal{S}_{pair} \leftarrow \{(t^{+},s^{+}),(t^{+},s^{-}_{1}) \dots (t^{+},s^{-}_{k})\}$ \;
    $\theta_s \leftarrow \theta_s + r\nabla\left(\mathcal{L}_{\mathcal{S}_{pair}}(h, w_t)\right)$ \;
}

 \caption{Guided Adversarial Contrastive Distillation (GACD)}
 \label{algo}
\end{algorithm}

\begin{table*}[h]
\resizebox{\textwidth}{!}{%
\begin{tabular}{ccccccccccccc}
\toprule
Teacher & \multicolumn{6}{c}{ResNet 18}                                                                       & \multicolumn{6}{c}{WRN-34-2}                                                                            \\ \cmidrule(r){2-7} \cmidrule(r){8-13}
Student & \multicolumn{2}{c}{ResNet 18}   & \multicolumn{2}{c}{WRN-34-2}         & \multicolumn{2}{c}{MobileNetV2} & \multicolumn{2}{c}{ResNet 18}  & \multicolumn{2}{c}{WRN-34-2}         & \multicolumn{2}{c}{MobileNetV2} \\ 
\midrule
Method  & Nat.           & Adv.           & Nat.           & Adv.           & Nat.           & Adv.           & Nat.          & Adv.           & Nat.           & Adv.           & Nat.           & Adv.           \\
KD      & 76.13          & 40.13          & 73.12          & 34.16          & 76.86          & 38.21          & \textbf{90.50} & 4.63           & \textbf{86.88} & 0.08           & \textbf{92.49} & 5.46           \\
ARD     & 79.49          & 51.21          & 79.05          & 50.16          & 79.47          & \textbf{50.22}          & 82.88         & 45.65          & 81.76          & 48.1           & 82.00             & 46.25          \\
GACD+AFT & \textbf{82.73} & \textbf{52.75} & \textbf{85.35} & \textbf{55.19} & \textbf{79.59} & 50.10  & 82.33         & \textbf{49.42} & 85.96 & \textbf{52.54} & 79.18          & \textbf{49.07} \\ \midrule
GACD    & 84.14          & 42.12 & 86.28          & 45.66 & 81.84          & 39.51 & 84.16         & 40.60  & 86.74          & 44.64 & 81.13          & 36.05 \\
\bottomrule
\end{tabular}%
}
\caption{Classification accuracy (\%) of student models with different methods on CIFAR-10. For each teacher, there are three students with different architecture style and network capacity. We mainly compare GACD with KD~\cite{hinton2015distilling} and ARD~\cite{DBLP:conf/aaai/GoldblumFFG20}. }
\label{tab:CIFAR-10}
\end{table*}

\begin{table*}[h]
\resizebox{\textwidth}{!}{%
\begin{tabular}{ccccccccccccc}
\toprule
Teacher & \multicolumn{6}{c}{ResNet 56}                                                                       & \multicolumn{6}{c}{WRN-40-2}                                                                        \\ \cmidrule(r){2-7} \cmidrule(r){8-13}
Student & \multicolumn{2}{c}{ResNet 32}   & \multicolumn{2}{c}{WRN-16-2}    & \multicolumn{2}{c}{MobileNetV2} & \multicolumn{2}{c}{ResNet 32}   & \multicolumn{2}{c}{WRN-16-2}    & \multicolumn{2}{c}{MobileNetV2} \\ \midrule
Method  & Nat.           & Adv.           & Nat.           & Adv.           & Nat.           & Adv.           & Nat.           & Adv.           & Nat.           & Adv.           & Nat.           & Adv.           \\
KD      & \textbf{60.69} & 4.94           & \textbf{63.51} & 4.95           & \textbf{55.93} & 4.70           & \textbf{60.80} & 2.99           & \textbf{62.67} & 3.68           & \textbf{56.15} & 3.69           \\
ARD     & 50.85          & 12.60          & 50.32          & 13.11          & 51.78          & 13.60          & 49.94          & 14.47          & 49.27          & 16.19          & 49.09          & \textbf{20.58} \\
GACD+AFT & 49.86          & \textbf{24.28} & 51.81          & \textbf{24.65} & 49.43          & \textbf{19.93} & 50.58          & \textbf{24.27} & 52.11          & \textbf{25.05} & 49.21          & 19.75          \\ \midrule
GACD    & 52.26          & 13.68          & 49.71          & 13.60          & 45.08          & 10.05          & 49.62          & 13.41          & 52.59          & 14.03          & 44.63          & 10.69  \\
\bottomrule
\end{tabular}%
}
\caption{Classification accuracy (\%) of student models with different methods on CIFAR-100. 
}
\label{tab:CIFAR-100}
\end{table*}

\section{Experimental Results}
In this section, we conduct a series of experiments to evaluate adversarial robustness and transferability of our method.
Then we illustrate the selection of hyper-parameters, and study the influence of teacher's error on distillation.

\subsection{Datasets}
Totally we consider three different datasets in our experiments: (1) CIFAR-10~\cite{krizhevsky2009learning} contains 60K images in 10 classes, of which 50K for training and 10K for testing. (2) CIFAR-100. It is just like CIFAR-10, except it has 100 classes containing 600 images each. (3) STL-10~\cite{coates2011analysis}. There is a training set of 50K labeled images from 10 classes and 100K unlabeled images, and a test set of 8K images.

We use CIFAR-10 and CIFAR-100 for evaluating classification accuracy of different distillation methods, while STL-10 and CIFAR-100 are used to test the transferability of distilled students.
\subsection{Implementation Details}
\subsubsection{Construction of Sample Set.} 
There are negatives and positives within a sample set.
As our method is supervised, we sample negatives from different classes rather than different instances, when picking up a positive sample from the same class.
However, we did some modifications on positive samples.
As suggested in~\cite{DBLP:conf/aaai/GoldblumFFG20}, not all robust models are good teachers, so that distillation with natural images often results unexpected failures.
In our view, adversarial examples are like hard examples supporting the decision boundaries.
Without hard examples, the distilled models would certainly make mistakes.
Thus, we adopt a self-supervised way to generate adversarial examples using Projected Gradient Descent (PGD).
Given a certain budget of perturbations, we aim to find the perturbation which leads to maximal distortion on features.
The distance metric we use of features is Wasserstein Distance used in~\cite{NIPS2019_8459}.

\subsubsection{Estimation Model $h$.}
$h$ is used in the binary classification problem so that we can estimate the distribution \mbox{$q(C=1 \mid t, s)$}.
The only requirement for $h$ is that the output of $h$ has to be in the range of $[0,1]$; for example,
\mbox{$h(t, s)=\frac{e^{t^{\prime} s / T}}{e^{t^{\prime} s / T}+\frac{k}{M}}$}, where $T$ is the temperature and $M$ is the cardinality of the dataset.
Alternatively, $h$ can be a network like the discriminator in Generative Adversarial Network (GAN)~\cite{goodfellow2014generative}.
We use the former in our experiments.

\subsubsection{Adversarial Fine-Tuning.}
Existing contrastive learning methods leverage \textit{linear evaluation} for downstream tasks~\cite{dosovitskiy2015discriminative,2020arXiv200205709C,2019arXiv191105722H}.
Concretely, it requires to learn a linear layer $l_{\psi}(\cdot)$ on top of the fixed and well-trained contrastive learning models $f_{\theta}(\cdot)$.
But it is proved in~\cite{2020arXiv200510190A} that deep models are not guaranteed to be adversarially robust if only low level features are robust.
In contrastive learning, last layer is vulnerable to adversarial examples.
So we do full network fine-tuning with adversarial training after feature distillation.

\subsection{Robustness Evaluation of Student Model}
\subsubsection{Setup.} We experiment on CIFAR-10 and CIFAR-100 with different student-teacher combinations of various model capacity.
For CIFAR-10, ResNet18~\cite{he2016deep} and WideResNet-34-10~\cite{DBLP:journals/corr/ZagoruykoK16} are teacher models, while for CIFAR-100 we use ResNet56 and WideResNet-40-2.
The performances of adversarially trained teacher models are listed in Table~\ref{tab:teacher}.
We use different teacher-student combinations to show our approach is model agnostic.
More details will be introduced in following sections.

\begin{table}[htb]
\resizebox{\columnwidth}{!}{
\begin{tabular}{ccccc}
\toprule
Dataset  & \multicolumn{2}{c}{CIFAR-10} & \multicolumn{2}{c}{CIFAR-100} \\ \cmidrule(r){2-3} \cmidrule(r){4-5} 
Model    & Resent18       & WRN-34-10        & ResNet56       & WRN-40-2         \\
\midrule
Nat.  & 76.54          & 84.41      & 59.29          & 60.27       \\
Adv.  & 44.46          & 45.75      & 20.32          & 22.40      \\
\bottomrule
\end{tabular}%
}
\caption{Classification accuracy (\%) of different adversarially trained teacher models on CIFAR-10 and CIFAR-100. We use Nat. to denote classification accuracy on natural images, and Adv. on adversarial images.}
\label{tab:teacher}
\end{table}

\subsubsection{Results.}

We mainly compare GACD with KD~\cite{hinton2015distilling} and ARD~\cite{DBLP:conf/aaai/GoldblumFFG20}.
Table~\ref{tab:CIFAR-10} and Table~\ref{tab:CIFAR-100} summarize the performances of different distillation methods on CIFAR-10 and CIFAR-100 from two aspects: natural accuracy and adversarial accuracy under 20-step PGD attack.
Unless specified, we use perturbation budget $\epsilon=8 / 255$ under $l_{\infty}$, and 20-step PGD attack as the default in our experiments.
We also investigate the influence of model architectures.
Specifically, we select three different students for each teacher.
These students differ from model capacity to architectural style.

We use ResNet18, WRN-34-2 and MobileNet V2 as student models on CIFAR-10. 
Our approach has the best results in natural accuracy and adversarial accuracy for most student-teacher combinations (see Table~\ref{tab:CIFAR-10}), which exceed teacher models as well.
While on CIFAR-100, we use ResNet32, WRN-40-2, and MobileNetV2 as student models.
As shown in Table~\ref{tab:CIFAR-100}, student models produced by our approach have the best adversarial robustness among all the methods.
Students with KD have best natural accuracy bust worst adversarial robustness.

\subsection{Transferability of Distilled Features} 
\subsubsection{Setup.}
In representation learning, a primary goal is to learn general knowledge.
In other words, the representations or features learned could be applied to different tasks or datasets that are not used for training.
Therefore, in this section we test if the features distilled transfer well.

We use ResNet18 as teacher and student models.
In our experiment, models are frozen once trained and used to extract features later (the layer prior to the logit).
Then we train a linear classifier on the top of frozen student models to perform classification, and show the transferability of the distilled features on STL-10 and CIFAR-100.

\subsubsection{Results.}
We compared GACD with different methods and the results are reported in Table~\ref{tab:transfer}.
As illustrated, all distillation methods improve the transferability of learned features on both natural images and adversarial examples.
Our method shows the best performances on both datasets with average 11.8\% improvement on natural accuracy and 5.84\% on adversarial accuracy.

\begin{table}[h]
\resizebox{\columnwidth}{!}{%
\begin{tabular}{cccccc}
\toprule
\multicolumn{2}{c}{Datasets}                                        & \multicolumn{2}{c}{STL10}                             & \multicolumn{2}{c}{CIFAR-100}                          \\
\cmidrule(r){1-2} \cmidrule(r){3-4} \cmidrule(r){5-6}
\multicolumn{2}{c}{Methods}             & Nat.                      & Adv.                      & Nat.                      & Adv.                      \\
\midrule
Teacher                  & \multicolumn{1}{l}{Adv. Training} & 48.81                     & 28.40                      & 27.17                     & 11.42                     \\
\multirow{3}{*}{Student} & KD                                & \multicolumn{1}{l}{56.79} & \multicolumn{1}{l}{29.33} & \multicolumn{1}{l}{30.64} & \multicolumn{1}{l}{12.03} \\
                         & ARD                               & 59.94                     & 35.16                     & 32.17                     & 15.06                     \\
                         & GACD (Ours)                              & \textbf{63.05}                     & \textbf{35.48}                     & \textbf{36.52}                     & \textbf{16.02}                    \\
\bottomrule
\end{tabular}%
}
\caption{Illustration of transferability of different student models. Here we use 7-step PGD attack to evaluate adversarial robustness (\%).}
\label{tab:transfer}
\end{table}

\subsection{Hyper-parameters}
We investigate the influence of two main hyper-parameters used in our method: (1) the number of negative samples $k$ in Eq.~\ref{eqn:loss h}, and (2) the temperature $T$ which suppresses softmax probability. 
The architectures of teacher and student are both ResNet18.
But note that our method is model agnostic.
Experiments are conducted on CIFAR-10 and the results are shown in Figure~\ref{fig:param}.
\subsubsection{Number of Negative Samples $k$.}
We validated a series of different $k$: 16, 64, 256, 1024, 4096 and 16384.
As shown in Figure~\ref{fig:param neg}, adversarial robustness increases as $k$ gets large.
Natural accuracy falls down due to the trade-off~\cite{pmlr-v97-zhang19p}.
As the accuracy is highest when $k=16384$, we use $k=16384$ in all experiments reported.
\subsubsection{Temperature.}
\begin{figure}[htb]
\centering
\begin{subfigure}{.25\textwidth}
  \centering
  \includegraphics[width=\linewidth]{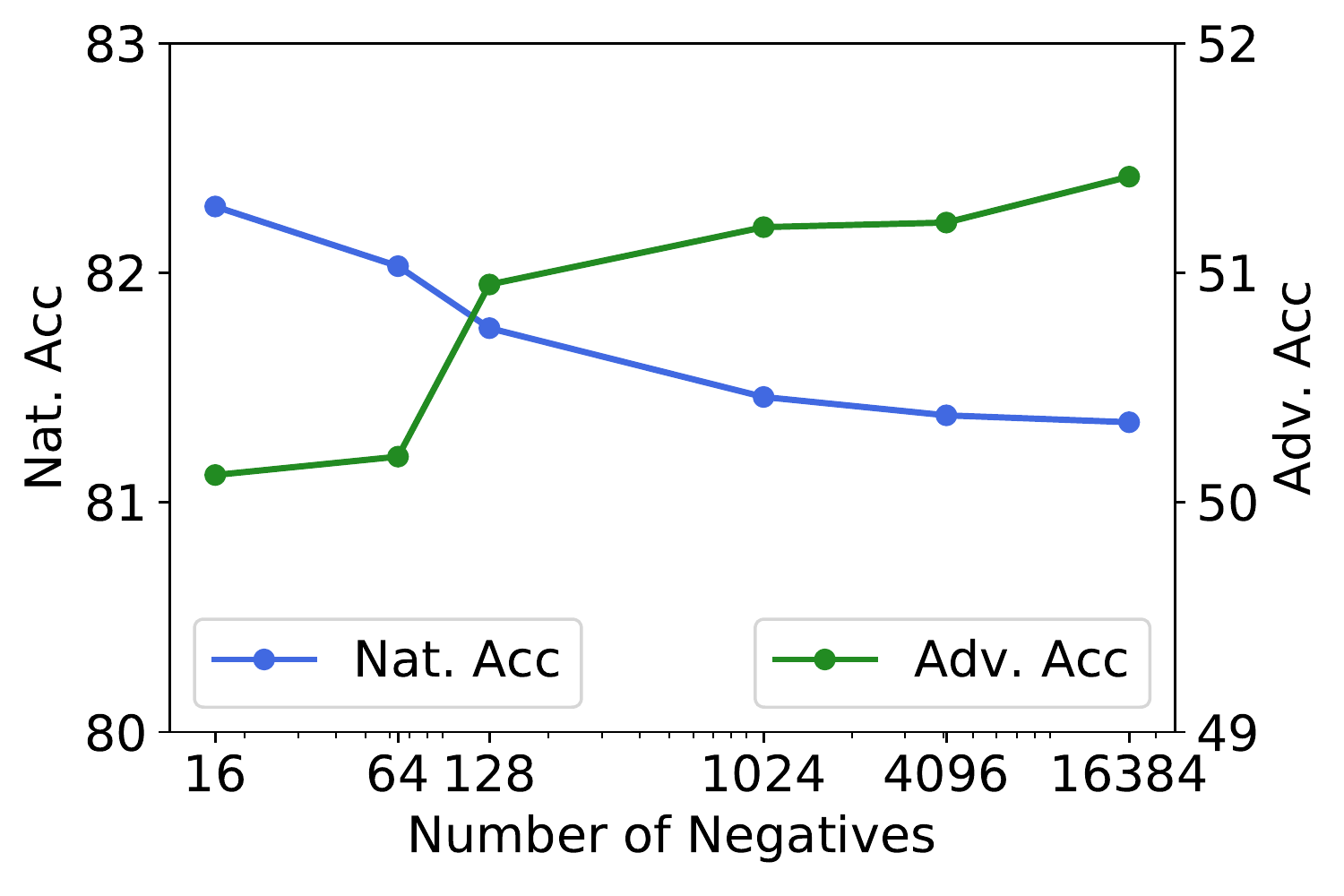}

  \caption{Effects of varying $k$.}
  \label{fig:param neg}
\end{subfigure}%
\begin{subfigure}{.25\textwidth}
  \centering
  \includegraphics[width=\linewidth]{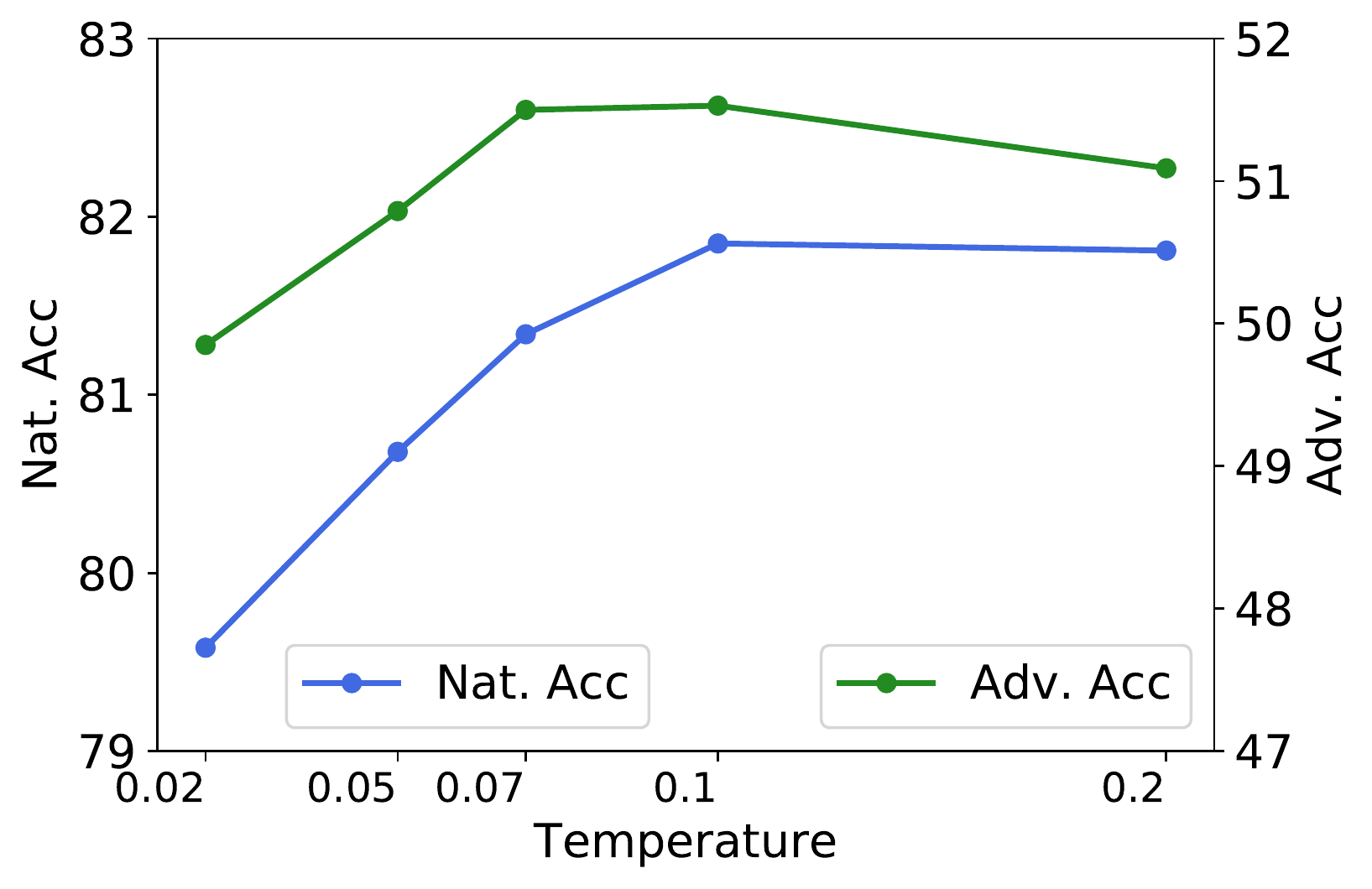}

  \caption{Effects of varying T.}
  \label{fig:param temp}
\end{subfigure}

\caption{Effects of varying the number of negative samples and the temperature.}
\label{fig:param}
\end{figure}

We experimented with Temperature $T$ between 0.01 and 0.3, and illustrate the results when $T=0.02,0.05,0.07,0.1, 0.2$ in Figure~\ref{fig:param temp}.
As we can see, the best natural and adversarial accuracy are obtained when $T=0.1$, while some extreme values give sub-optimal solutions.
Thus, we picked up $T=0.1$ in our experiments.

\subsection{Ablation Study}
In GACD, we assign small importance on samples which are misclassified by teacher.
The reason is that we believe wrong predictions made by the teacher would have a negative influence on students.
For illustration, we temporarily remove sample importance in GACD to see the difference.
We select ResNet18 as the teacher while WRN-34-10 as the student, and validate on CIFAR-10.
The natural and adversarial accuracy we got are $83.53\%$ and $54.23\%$ for WRN-34-10.
Apparently there is a drop compared to the original (see Table~\ref{tab:CIFAR-10}).

In addition, we applied sample re-weighting to ARD to evaluate its effectiveness.
Other settings or parameters in ARD are exactly the same.
With ResNet18 as the teacher, the classification results of student WRN-34-10 are 80.59\% and 54.23\%, increased by 1.54\% and 4.07\% on natural accuracy and adversarial accuracy respectively. 

So we can conclude sample re-weighting indeed helps improve the performances of students in distillation.

\section{Distilled Feature Analysis}
In this section, we provide a detailed analysis of latent representations or features extracted by students with different methods.

\subsection{Inter-class Correlations}
For classification problems, cross-entropy is widely used as the objective function.
It, however, ignores the correlations between classes.
Knowledge distillation solves this problem with a teacher model.
Soft labels are the key component for its success, which inherently contains correlations between classes.
Such correlations contribute to the performances of student models.
To illustrate the capability of capturing correlations in different methods, we computed the differences between the correlation metrices of the teachers' and students' logits.

\begin{figure}[h]
\centering
\begin{subfigure}{.25\textwidth}
  \centering
  \includegraphics[width=\linewidth]{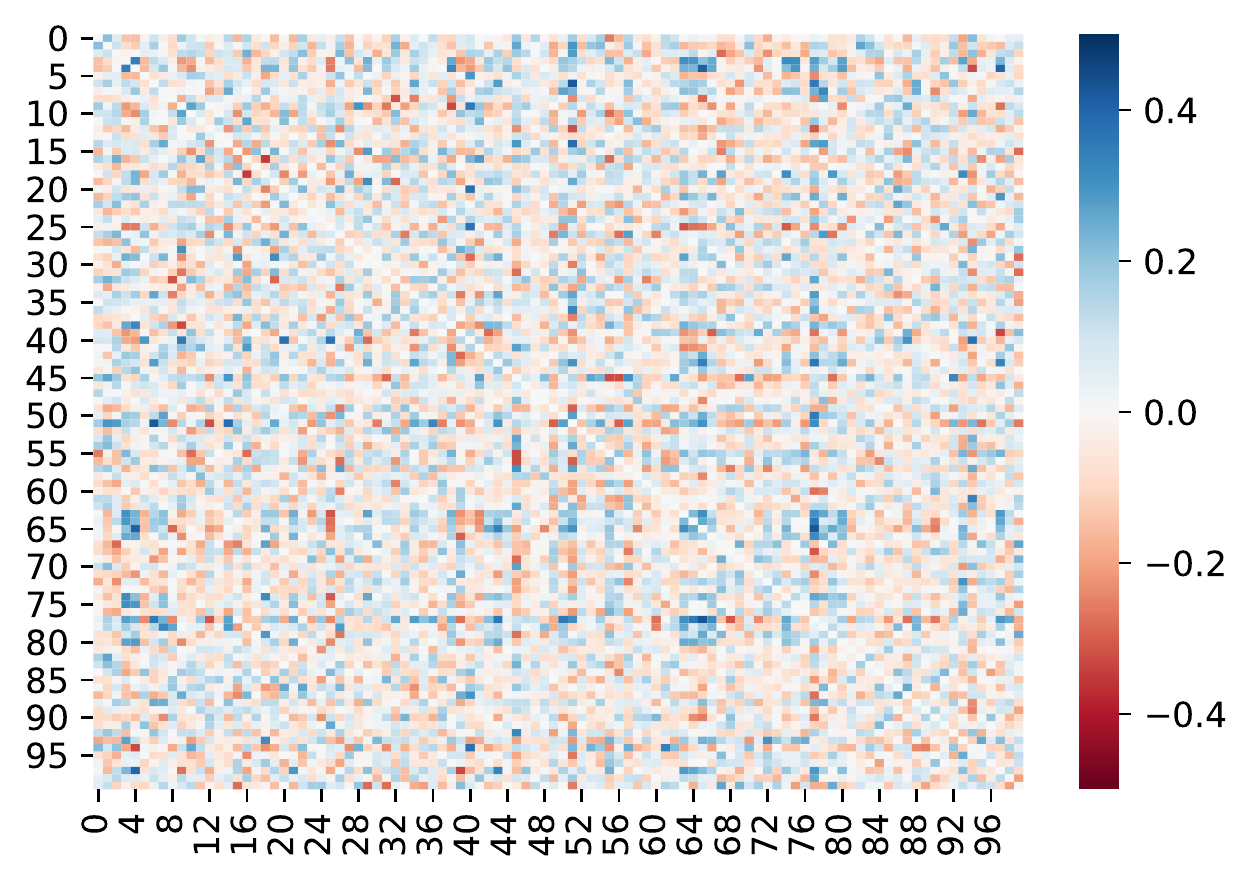}
  \caption{Student: ARD}
  \label{fig:corre ard cln}
\end{subfigure}%
\begin{subfigure}{.25\textwidth}
  \centering
  \includegraphics[width=\linewidth]{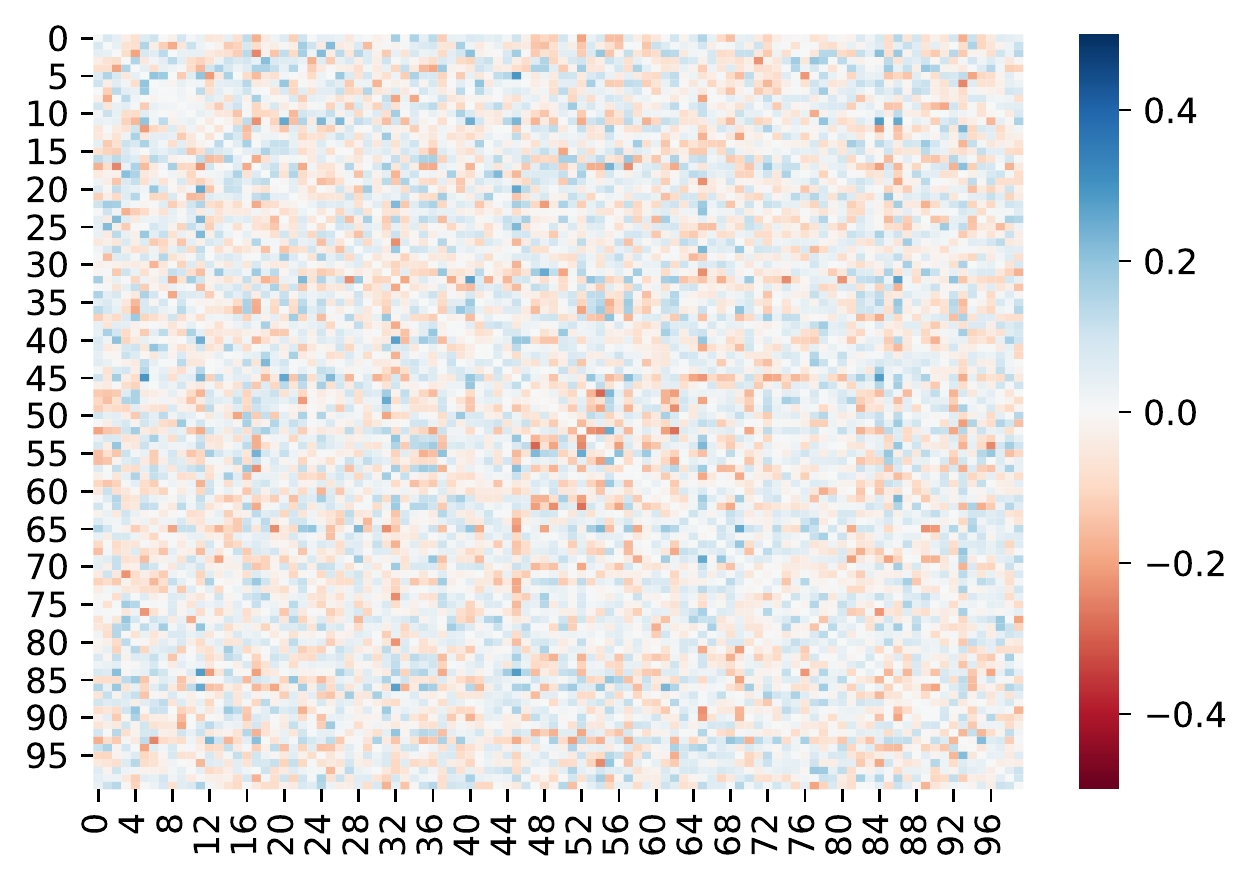}
  \caption{Student: GACD (ours)}
  \label{fig:corre gacd cln}
\end{subfigure}

\caption{Differences of logits correlations between teachers and students on natural data from CIFAR-100. For visualization, we use WRN-40-2 as teacher and WRN-16-2 as student.}
\label{fig:corre cln}
\end{figure}

\begin{figure}[h]
\centering
\begin{subfigure}{.25\textwidth}
  \centering
  \includegraphics[width=\linewidth]{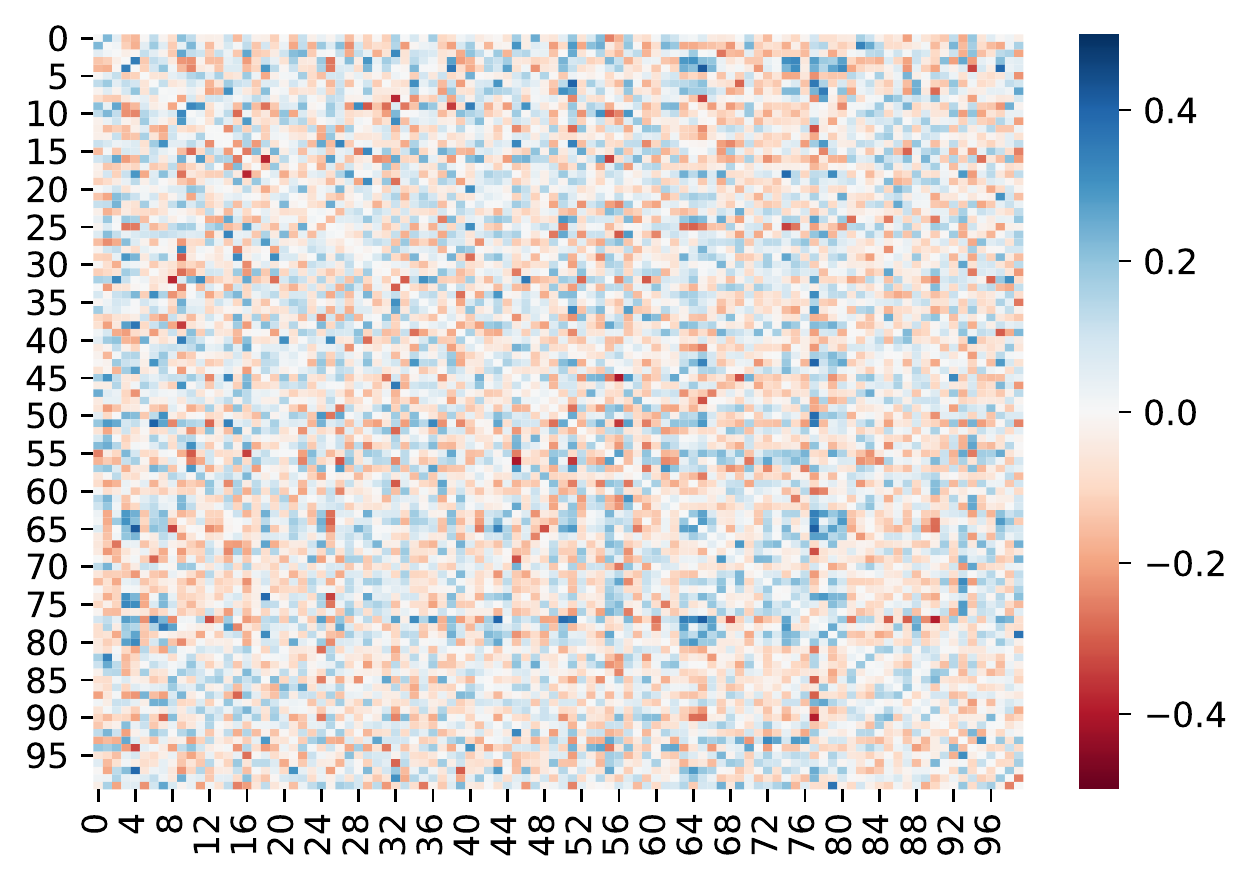}
  \caption{Student: ARD}
  \label{fig:corre ard adv}
\end{subfigure}%
\begin{subfigure}{.25\textwidth}
  \centering
  \includegraphics[width=\linewidth]{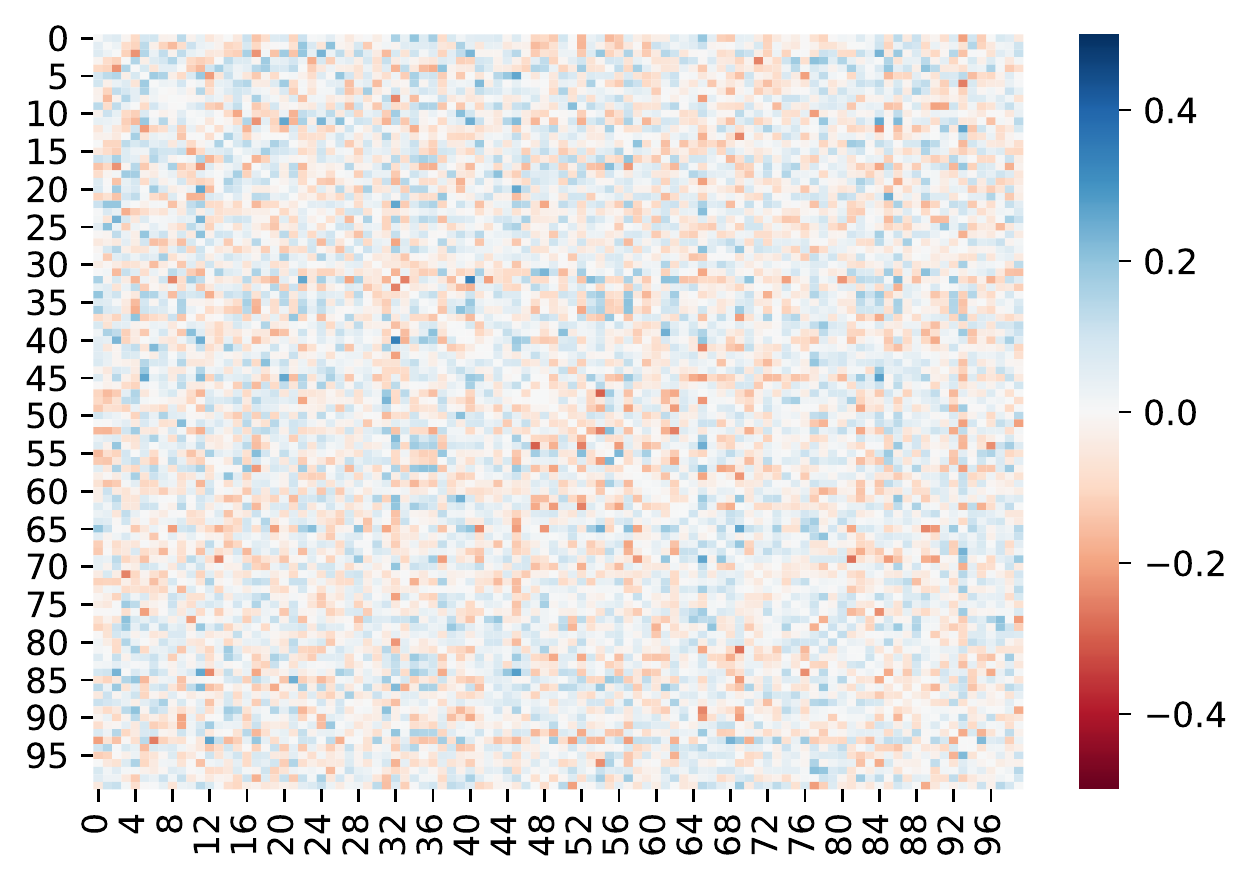}
  \caption{Student: GACD (ours)}
  \label{fig:corre gacd adv}
\end{subfigure}

\caption{Differences of logits correlations between teachers and students on adversarial data from CIFAR-100. Same models are used as in Figure~\ref{fig:corre cln}.}
\label{fig:corre adv}
\end{figure}

Since the objective functions of KD and ARD are highly similar, we select ARD and compare with our method.
\mbox{Figure~\ref{fig:corre cln}} shows the differences with natural images as inputs, while Figure~\ref{fig:corre adv} shows the differences on adversarial images.
Clearly we can see there are significant reductions (light color) of differences between teachers and students with our method, compared to ARD.
This means our method captures more structural knowledge during distillation, which is also supported by the increased accuracy. 

\subsection{Feature under Attacks}
\begin{figure}[h]
\centering
\begin{subfigure}{.49\columnwidth}
  \centering
  \includegraphics[width=\textwidth]{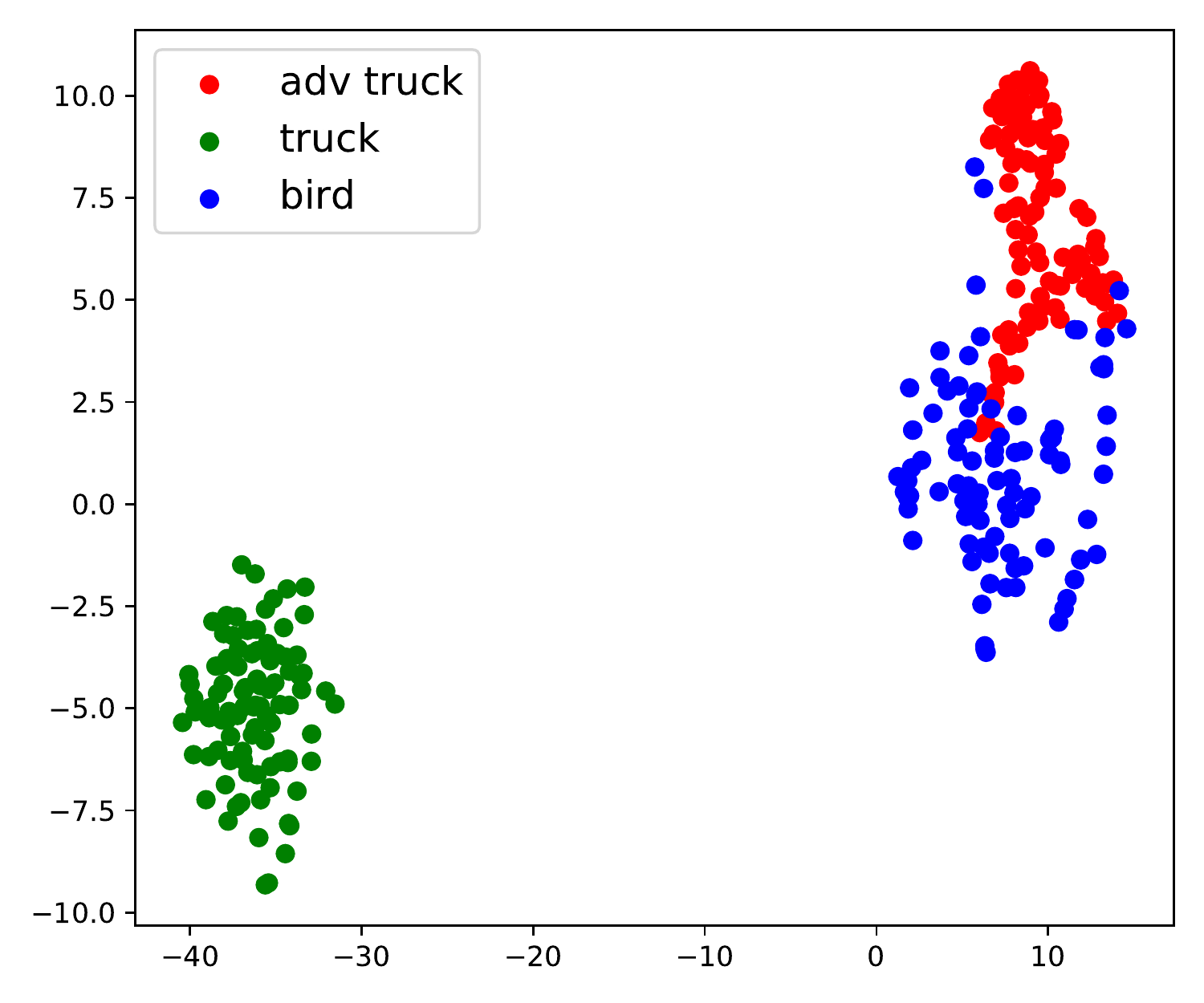}
  \caption{Standard undefended model.}
  \label{fig:feat stu nat}
\end{subfigure}%
\begin{subfigure}{.49\columnwidth}
  \centering
  \includegraphics[width=\textwidth]{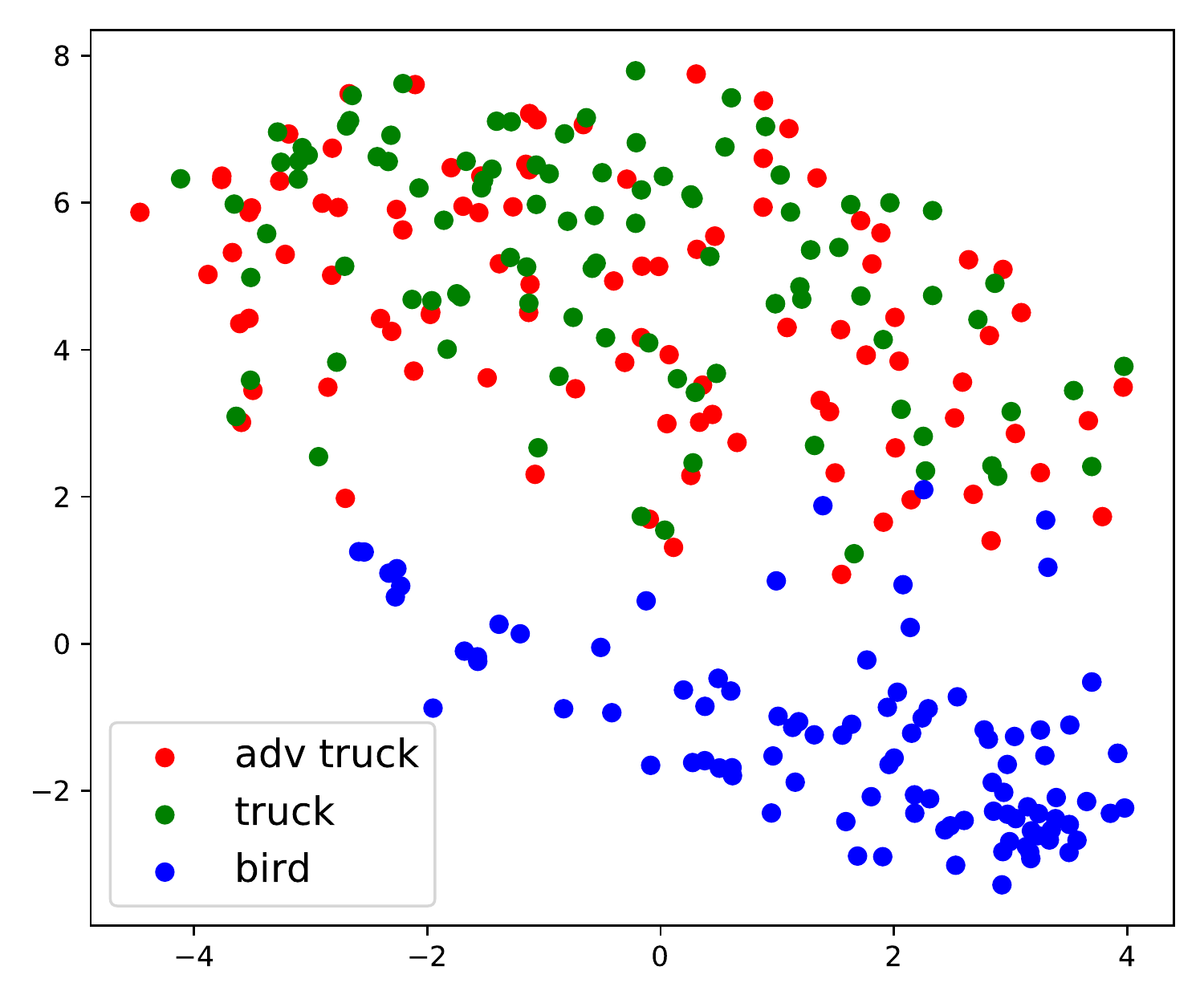}
  \caption{Teacher.}
  \label{fig:feat teacher}
\end{subfigure}

\begin{subfigure}{.49\columnwidth}
  \centering
  \includegraphics[width=\textwidth]{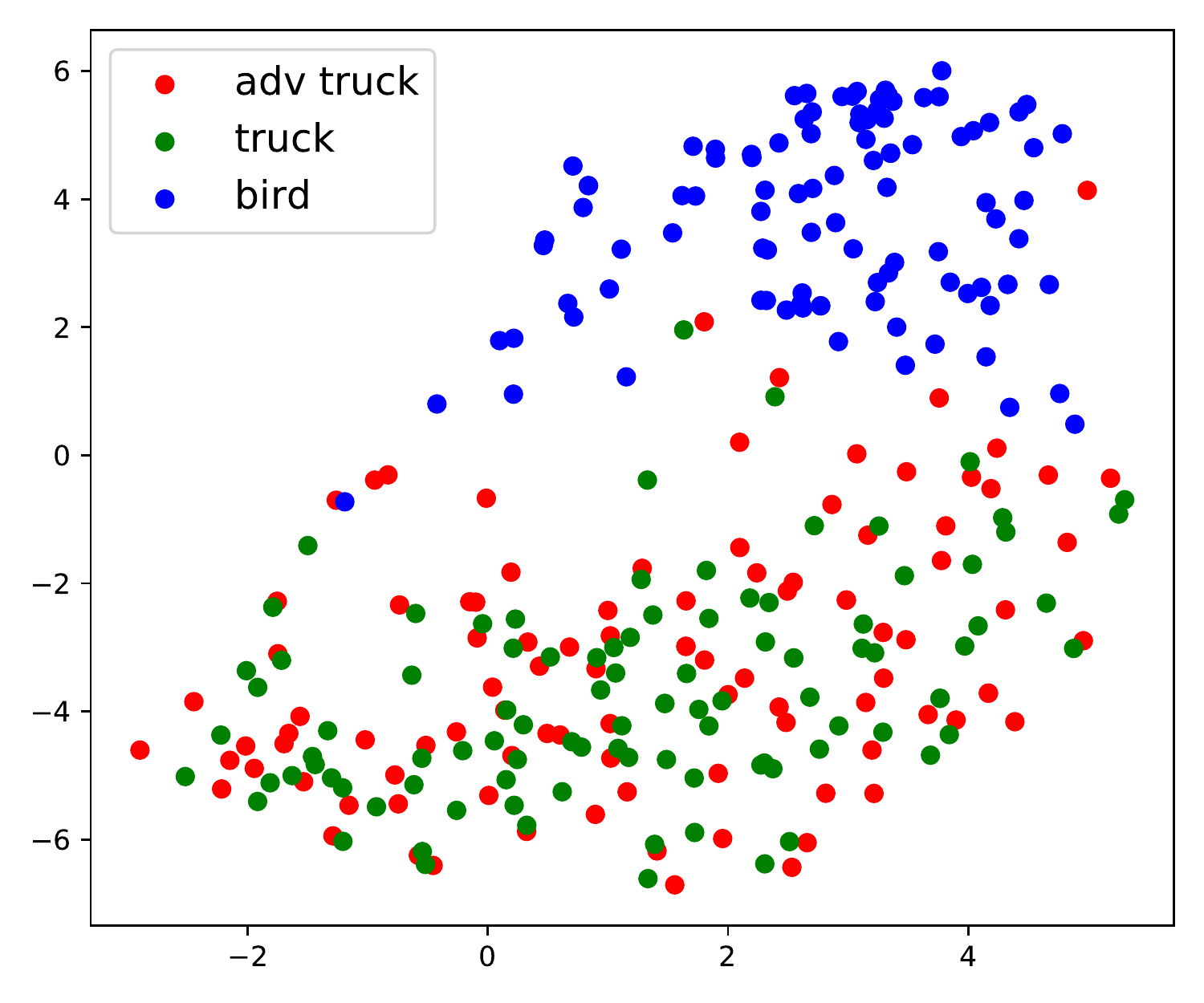}
  \caption{Student: ARD}
  \label{fig:feat stu ard}
\end{subfigure}
\begin{subfigure}{.49\columnwidth}
  \centering
  \includegraphics[width=\textwidth]{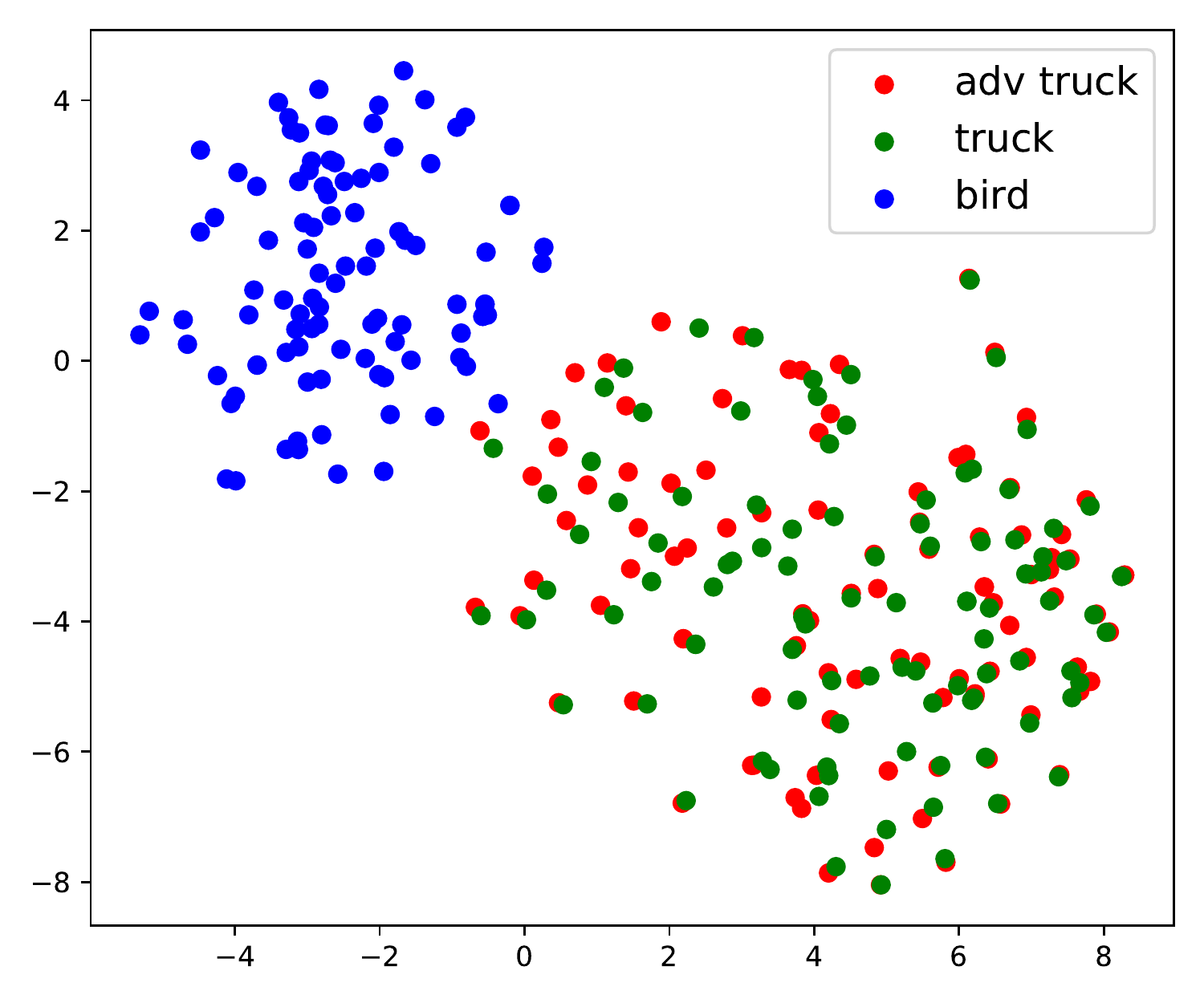}
  \caption{Student: GACD (ours)}
  \label{fig:feat stu ours}
\end{subfigure}

\caption{Illustration of latent representation generated by different models. The \textit{blue} and \textit{green} points are 100 randomly sampled natural images from class 'bird' and 'truck' respectively, while the \textit{red} points are adversarial images crafted from images from class truck.}
\label{fig:feat vis}
\end{figure}

We further investigate how natural images and adversarial images are represented in feature space by different models.
The t-SNE visualization of high dimensional latent representations of sampled images is shown in Figure~\ref{fig:feat vis}.
Concretely, we sampled two-class images (bird and truck) from CIFAR-10 for illustration, and crafted adversarial images (namely adv truck) which originally belong to truck but misclassified as bird.
The \textit{green} and \textit{blue} points indicate the natural images of trucks and birds, while \textit{red} points represent images of adv trucks.
Then these samples are fed into four models: standard undefended model, adversarially trained teacher, and two student models with ARD and our method GACD.
For the standard undefended model (Figure~\ref{fig:feat stu nat}), all samples of adv trucks are misclassified birds (red points are mixed with blue points), and far from the original class (green points).
Other three models show adversarial robustness as most samples from adv truck as classified as trucks.
However, there are several red points falling into the green cluster for student with ARD (Figure~\ref{fig:feat stu ard}). 
Besides, there is no clear boundary as data points are kind of mixed.
In contrast, with our proposed GACD, Figure~\ref{fig:feat stu ours} clearly shows larger distances between classes and smaller intra-class distances.
The differences in feature space are also reflected on adversarial classification accuracy.




\section{Conclusion}
In this paper, we present a novel approach: Guided Adversarial Contrastive Distillation (GACD) to transfer adversarial robustness with features, which is different from existing distillation methods.
Theoretically, we formulate our distillation problem into contrastive learning, and connect it to mutual information. 
Taking teacher's error into consideration, we propose sample reweighted noise contrastive estimation, which is proved to be applicable to other distillation methods as well.
Compared to other methods in extensive experiments, Our method captures more structural knowledge and shows comparable or even better performances.
In addition, our method has the best transferability across tasks or models. 
In the future, we will look deep into deep learning models, find the key property leading to adversarial robustness and develop efficient methods for distillation.

\bibliography{references.bib}

\end{document}